\documentclass[10pt,twocolumn,letterpaper]{article}

\usepackage{soul}
\usepackage{color,xcolor}
\usepackage{graphicx}
\usepackage{amsmath}
\usepackage{amssymb}
\usepackage{booktabs}
\usepackage{multirow}
\usepackage[accsupp]{axessibility}
\sethlcolor{yellow}
\usepackage[pagenumbers]{config/cvpr}

\usepackage{times}
\usepackage{epsfig}
\usepackage{graphicx}
\usepackage{amsmath}
\usepackage{amssymb}
\usepackage{caption}
\usepackage{enumitem}
\usepackage{booktabs}

\usepackage{caption}
\usepackage{makecell}
\usepackage{multirow}
\usepackage{xcolor, colortbl}
\usepackage{cuted}
\usepackage{capt-of}
\usepackage{comment}
\usepackage[normalem]{ulem} %
\usepackage{pifont}

\usepackage[pagebackref=true,breaklinks=true,colorlinks,bookmarks=false,pagebackref]{hyperref}

\usepackage[normalem]{ulem}

\usepackage{afterpage}

\newcommand{\modelname}{\mbox{TokenHMR}\xspace}

\newcommand{\smpl}{\mbox{SMPL}\xspace}

\newcommand{\hps}{\mbox{HPS}\xspace}

\newcommand{\cheading}[1]{\noindent\textbf{#1.}}

\usepackage{lipsum}
\usepackage{balance}

\definecolor{citecolor}{HTML}{0071bc}
\definecolor{frontcolor}{HTML}{325ea5}
\definecolor{sidecolor}{HTML}{a58b77}
\definecolor{DeltaColor}{rgb}{0.039,0.73,0.71}
\definecolor{SigmaColor}{rgb}{0.98,0.45,0.0}
\definecolor{AlphaColor}{rgb}{0,0,0.8}
\definecolor{BetaColor}{rgb}{0.8,0,0.8}
\definecolor{GammaColor}{rgb}{0.514,0.34,0.224}
\definecolor{EpsilonColor}{rgb}{0.353,0.725,0.906}
\definecolor{PurpleColor}{HTML}{bca5ea}
\definecolor{OrangeColor}{rgb}{0.914,0.541,0.0.141}
\definecolor{GreenColor}{rgb}{0.137,0.573,0.565}
\definecolor{RedColor}{rgb}{0.949,0.275, 0.224}
\definecolor{LightCyan}{rgb}{0.88,1,1}
\definecolor{Gray}{gray}{0.85}

\newcommand{\dataColor}{black}

\newcommand{\threedpw}{\mbox{\textcolor{\dataColor}{3DPW}}\xspace}
\newcommand{\threeDPW}{\threedpw}

\newcommand{\amass}{\mbox{AMASS}\xspace}
\newcommand{\moyo}{\mbox{MOYO}\xspace}

\newcommand{\bedlam}{\mbox{BEDLAM}\xspace}

\newcolumntype{a}{>{\columncolor{Gray}}c}

\newcommand{\REFINE}[1]{{\color{black} #1}}
\newcommand{\CR}[1]{{\color{black} #1}}
\newcommand{\moveToSupMat}[1]{\begin{comment}#1\end{commment}}
\newcommand{\supmat}{\textcolor{magenta}{\emph{Sup.~Mat.}}\xspace}

\newcommand{\vposer}{\mbox{VPoser}\xspace}

\newcommand{\colorRef}[1]{\textcolor{red}{#1}} %
\usepackage[capitalize]{cleveref}
\crefname{figure}{\colorRef{Fig.}}{\colorRef{Figs.}}
\Crefname{figure}{\colorRef{Figure}}{\colorRef{Figures}}
\crefname{section}{\colorRef{Sec.}}{\colorRef{Secs.}}
\Crefname{section}{\colorRef{Section}}{\colorRef{Sections}}
\Crefname{table}{\colorRef{Table}}{\colorRef{Tables}}
\crefname{table}{\colorRef{Tab.}}{\colorRef{Tabs.}}
\Crefname{equation}{\colorRef{Equation}}{\colorRef{Equations}}
\crefname{equation}{\colorRef{Eq.}}{\colorRef{Eqs.}}

\renewcommand{\ie}{\mbox{i.e.}\xspace}

\newcommand{\groundtruth}{\mbox{ground-truth}\xspace}

\newcommand{\twoD}{2D\xspace}
\newcommand{\threeD}{3D\xspace}

\newcommand{\vqvae}{\mbox{VQ-VAE}\xspace}
\newcommand{\tokenizer}{tokenizer\xspace}
\newcommand{\codebook}{\mbox{codebook}\xspace}
\newcommand{\losscutShort}{\mbox{TALS}\xspace}
\newcommand{\losscutFull}{\mbox{Threshold-Adaptive Loss Scaling}\xspace}

\newcommand{\embedding}{e\xspace}
\newcommand{\codebookSmall}{c\xspace}
\newcommand{\codebookBig}{\mathit{CB}\xspace}
\newcommand{\pseudoGT}{p-GT\xspace}
\newcommand{\hmrTwo}{\mbox{HMR2.0}\xspace}

\newcommand{\regPose}{\theta}
\newcommand{\gtPose}{\theta_g}

\newcommand{\regShape}{\beta}
\newcommand{\gtShape}{\beta_g}

\newcommand{\cam}{T}

\newcommand{\threeDJnts}{J_{3D}}
\newcommand{\twoDJnts}{J_{2D}}
\newcommand{\gtThreeDJnts}{J_{3D_g}}
\newcommand{\gtTwoDJnts}{J_{2D_g}}

\newcommand{\mesh}{\mathcal{M}}
\newcommand{\vertices}{V}

\newcommand{\image}{I}

\begin{document}

\title{TokenHMR: Advancing Human Mesh Recovery with \\a Tokenized Pose Representation}

\author{
    Sai Kumar Dwivedi$^{1,}$\thanks{${\ast}$ Joint first author} \quad 
    Yu Sun$^{2,\ast}$ \quad
    Priyanka Patel$^{1}$ \quad 
    Yao Feng$^{1,2,3}$ \quad
    Michael J. Black$^1$ \quad \\
    \normalsize $^1$Max Planck Institute for Intelligent Systems, T\"{u}bingen, Germany \quad
    \normalsize $^2$Meshcapade \quad
    \normalsize $^3$ETH Zurich
}

\twocolumn[{%
\renewcommand\twocolumn[1][]{#1}%
\maketitle
 \vspace*{-1cm}
\begin{center}
    \centering
    \captionsetup{type=figure}
    \includegraphics[width=1.\linewidth]{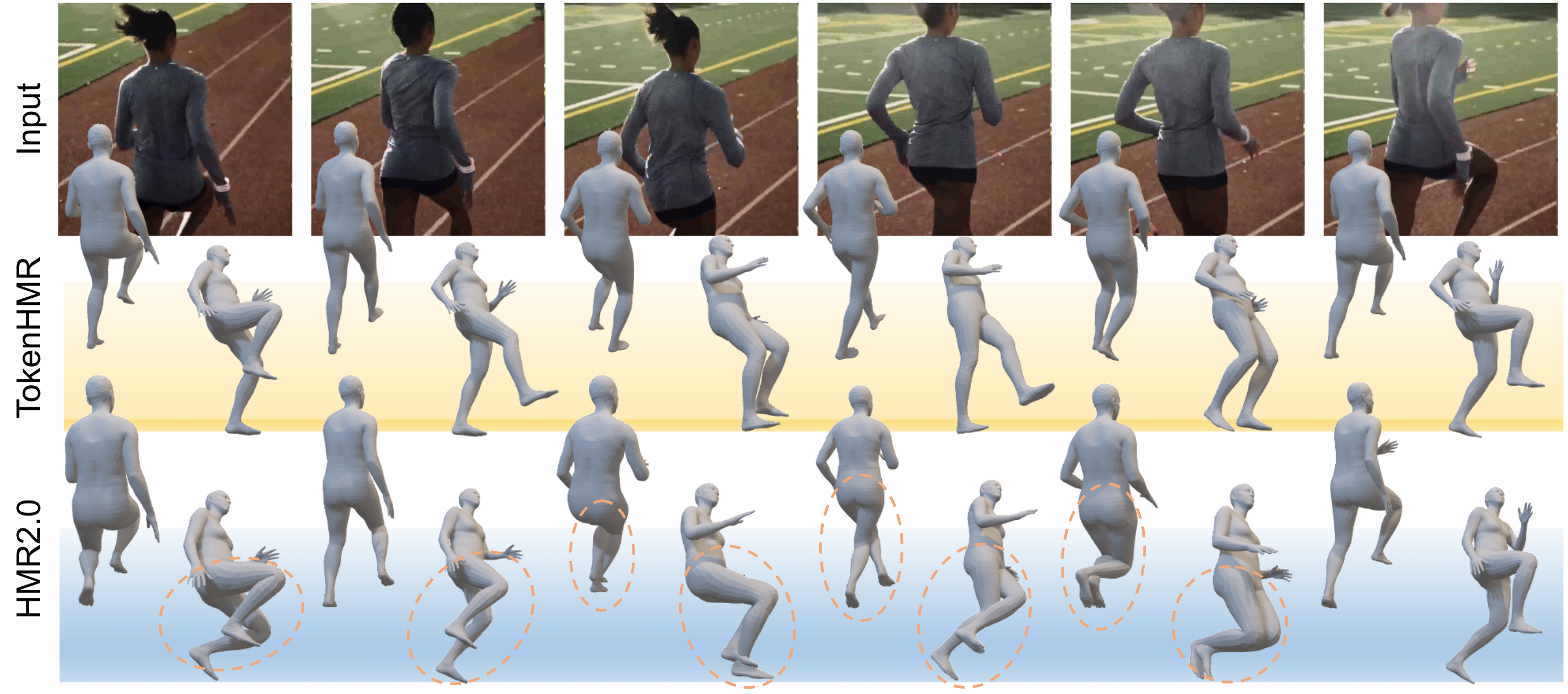}
    \captionof{figure}{Existing methods that regress 3D human pose and shape (HPS) from an image (like HMR2.0 \cite{hmr2}) estimate bodies that are either image-aligned or have accurate 3D pose, but not both. We show that this is a fundamental trade-off for existing methods. To address this our method, \modelname, introduces a novel loss, \textit{\losscutFull (\losscutShort)}, and a discrete token-based pose representation of 3D pose. With these, \modelname achieves state-of-the-art accuracy on multiple in-the-wild 3D benchmarks.
    }
\label{fig:teaser}
  \vspace*{0.3cm}
\end{center}%
}]

\def\thefootnote{*}\footnotetext{Equal contribution}

\begin{abstract}

We address the problem of regressing 3D human pose and shape from a single image, with a focus on 3D accuracy.
The current best methods leverage large datasets of 3D pseudo-ground-truth (p-GT) and 2D keypoints, leading to robust performance.
With such methods, however, we observe a paradoxical decline in 3D pose accuracy with increasing 2D accuracy.
This is caused by biases in the p-GT and the use of an approximate camera projection model.
We quantify the error induced by current camera models and show that fitting 2D keypoints and p-GT accurately causes incorrect 3D poses.
Our analysis defines the invalid distances within which minimizing 2D and p-GT losses is detrimental.
We use this to formulate a new loss, ``\losscutFull'' (\losscutShort), that penalizes gross 2D and p-GT errors but not smaller ones.
With such a loss, there are many 3D poses that could equally explain the 2D evidence.
To reduce this ambiguity we need a prior over valid human poses but such priors can introduce unwanted bias.
To address this, we exploit a tokenized representation of human pose and reformulate the problem as token prediction.
This restricts the estimated poses to the space of valid poses, effectively improving robustness to occlusion.
Extensive experiments on the EMDB and 3DPW datasets show that our reformulated loss and tokenization allows us to train on in-the-wild data while improving 3D accuracy over the state-of-the-art.
Our models and code are available for research at \href{https://tokenhmr.is.tue.mpg.de}{\tt{https://tokenhmr.is.tue.mpg.de}}.

\end{abstract}
    
\section{Introduction}
\label{sec:intro}

We address the problem of regressing \threeD human pose and shape (\hps) from a single image. %
Recent methods~\cite{hmr2,cliff,BEV,bedlam} are increasingly accurate on this task.
By accurate, we mean two things. A method should correctly regress the 3D pose but it should also align with the image evidence.
Unfortunately, current models cannot do both.
We observe a seeming paradox, that the more accurate a method is on fitting 2D keypoints, the less accurate it is at predicting 3D pose.
We study this problem and identify the common weak-perspective camera assumption as a key culprit.
This camera model does not match the true camera used to acquire the images and thus there is a mismatch between the projected 3D joints and the detected 2D ones.
\CR{
Since currently, no reliable method exists to estimate camera parameters from single image,}
we study and quantify this effect and propose two solutions to address it.

Specifically, we use the synthetic BEDLAM \cite{bedlam} dataset, which has perfect 3D and 2D groundtruth (GT).
We project the 3D data into 2D using the camera model from \cite{hmr2} to quantify the 2D error in the best case; it is large.
We then also go the other direction and show that low 2D error can result in large 3D error.
Even using a full perspective model like \cite{cliff} does not solve the problem since we lack the precise intrinsic and extrinsic camera parameters.

\CR{This analysis highlights the issue of supervising 3D pose regression with a 2D keypoint loss.}
\CR{
But such a loss opens up access to large datasets, providing generalization and robustness.}
Unfortunately,  pseudo ground-truth (p-GT) training data suffers the same problem since it is generated by fitting a 3D body to 2D data via optimization using an approximate camera model.
How can we leverage the abundant  information present in large-scale, in-the-wild, datasets while mitigating the decline in \threeD accuracy?
Our answer to this is {\em TokenHMR}, a new HPS regression method that strikes a balance between effectively leveraging \twoD keypoints while maintaining \threeD pose accuracy,
\CR{
thus leveraging Internet data today without known camera parameters.
}

TokenHMR has two main components.
The first is based on our key insight that supervision from \twoD keypoints, \CR{while flawed}, is valuable for preventing \CR{highly} incorrect predictions.
However, excessive reliance on 2D cues introduces bias.
\CR{To address this}, we define a new loss called \textit{\losscutFull (\losscutShort)} that penalizes large 
\REFINE{2D and p-GT errors} but only minimally penalizes small ones.
We use our BEDLAM analysis to define this, so that the network is not encouraged to fit 2D keypoints more accurately than makes sense given the camera model.

This, however, creates a new problem.
Predicting 3D pose from 2D keypoints is fundamentally ambiguous.
When one relaxes the keypoint matching constraint, even more 3D poses are consistent with the 2D data.
To control this, we need to introduce a prior that biases the network to {\em valid} poses.
Unfortunately, existing pose priors based on mixtures of Gaussians \cite{smplify} or VAEs \cite{smplx} are biased towards poses that occur frequently in the training data.
Instead, we seek an unbiased prior that restricts the network to only output valid poses but does not bias it to any particular pose.

This leads us to the second key component of TokenHMR, which gives it its name.
Specifically, we convert the problem of continuous pose regression into a problem of token prediction by tokenizing human poses.
We use a Vector Quantized-VAE (VQ-VAE) \cite{vqvae} to discretize continuous human poses by pre-training on extensive motion capture datasets, such as AMASS~\cite{amass} and MOYO~\cite{moyo}.
This tokenized representation provides the regressor with a ``vocabulary" of valid poses, effectively representing the the pose prior as a knowledge bank, {\em codebook}. 
Since VQ-VAE's are designed to represent a uniform prior, we posit that this reduces the biases caused by previous pose priors.

\modelname generates discrete tokens through classification, in contrast to regressing continuous pose.
When we take a SOTA HPS method and replace the continuous pose with our tokenized pose approach we see consistent improvements in 3D accuracy (all else held the same).

We perform extensive experiments to evaluate different ways of tokenizing pose and their effects on accuracy.
Any discretization of pose comes with some loss in accuracy. 
In our case it results in a loss of 3D accuracy of about 2.5mm, which is 20 times smaller than the accuracy of the state-of-the-art (SOTA) HPS regressors on real data; i.e., the loss in accuracy due to tokenization is negligible.

Finally, we put our two components together and find that they work synergistically.
Our new %
loss does not distort the 3D pose to over-fit the keypoints or p-GT and the tokenization keeps the network from distorting 3D pose for the sake of 2D accuracy.
With this combination, we achieve a new state-of-the-art in terms of 3D accuracy. %
We extensively evaluate \modelname and other recent methods on EMDB~\cite{emdb} and \threeDPW~\cite{threedpw}, which have accurate \threeD ground truth.
Using the same data and backbone, \modelname exhibits a 7.6\% reduction in 3D error compared to HMR2.0~\cite{hmr2} on the challenging EMDB dataset. 
Qualitative results suggest that the \modelname is robust to ambiguous image evidence and the estimated poses do not suffer from the ``bent knees'' bias of methods that use p-GT and 2D keypoints (see Fig.~\ref{fig:teaser}). 

In summary, we make the following key contributions:
(1) \textit{Analysis of \threeD Accuracy Degradation:} 
We analyze and quantify the trade-off between \threeD and \twoD accuracy that current \hps methods face if they use 2D losses.
(2) \textit{\losscutFull:} To ameliorate the issue, we develop a novel loss function that reduces the influence of \REFINE{2D and p-GT errors} that are less than the expected error due to the incorrect camera model.
(3) \textit{Token-Based Pose Representation:} We introduce a token-based representation for human pose and show that it produces more accurate pose estimates.
Our models and code are available for research.

\section{Related Work}
\label{sec:related_work}

\subsection{HPS Regression}
Estimating 3D human pose and shape from single images has been studied in great detail from optimization-based approaches to the most recent transformer-based regressors. 
Optimization approaches fit a parametric model~\cite{smpl, smplx, xu2020ghum} to 
 2D image cues, including, but not limited to keypoints~\cite{smplify, xu2020ghum, smplx}, silhouettes~\cite{nbf}, and part segmentations~\cite{unitepeople}. 
 Some learning-based approaches directly estimate the parametric body model from images~\cite{hmr, dsr, pare, BEV, poco, cliff, ROMP,pymaf,3dcrowdnet} and videos~\cite{vibe, humanMotionKanazawa19} and some estimate bodies with a model-free approach either as vertices~\cite{meshgraphormer,cmr,metro} or as implicit shapes~\cite{pifuhd,Mihajlovic_CVPR_2022,xiu2023econ}. 
Recent methods~\cite{hmr2, lin2023osx} use transformers to estimate 3D bodies, achieving the current best accuracy.
To address the challenges of generalization, recent methods like EFT~\cite{eft}, NeuralAnnot~\cite{NeuralAnnot}, HMR2.0~\cite{hmr2} and CLIFF~\cite{cliff}  use \REFINE{2D keypoints and p-GT} in the training loss, to produce a good alignment between the projected body and the image.
\CR{Methods like HuManiFlow~\cite{humaniflow} and POCO~\cite{poco} model probabilistic \hps to explicitly address pose ambiguity.}
The problem of 3D accuracy degradation in pursuit of better 2D alignment has been noted but not extensively quantified before our work. Our statistical analysis highlights this bias in existing HMR methods, offering a new perspective on training strategies for this problem.

Some methods~\cite{spec, cliff, zolly} address the 3D-to-2D projection error by estimating the camera from a single image.  
SPEC~\cite{spec} uses a network to predict camera parameters but does not generalize well, while CLIFF~\cite{cliff} uses an approximation by providing the network with information about the bounding box coordinates of  the person in the image. 
Estimating the camera from a single image is highly ill-posed so this remains a challenging, unsolved, problem.
Our approach reduces the impact of using the wrong camera model and can be applied to any \hps regression method.

\subsection{Pose Prior}
Human pose priors play a pivotal role in various applications like lifting 2D pose to 3D~\cite{smplify, smplx} and estimating human pose from images/videos~\cite{eft,prohmr}. 
Early pose priors focus on learning joint limits~\cite{joint_limits}  to avoid poses that are impossible.
Gaussian Mixture Models (GMMs) \cite{smplify}
and Generative Adversarial Networks (GANs)~\cite{hmr,georgakis2020hierarchical} are also used to impose prior knowledge during training. 
Some recent methods use VAEs~\cite{smplx} and normalizing flows~\cite{prohmr} as priors. 
Many of these methods are biased to commonly occurring poses and this bias is passed on the regressor. 
Methods like Pose-NDF~\cite{posendf} learn a manifold of plausible poses represented as the zero-level set of a neural implicit function. %
The mapping of invalid to valid poses involves gradient descent, which is an expensive operation when integrated in \hps training. 
In contrast to prior work, we learn a discrete token-based  prior over valid SMPL poses, reducing pose bias and improving robustness to occlusion, while being easy to integrate into \hps training.

We use a VQ-VAE~\cite{vqvae}, which is a variant of VAEs, to learn a discretized prior by quantizing the 3D training poses in a process called ``Tokenization'' creating a knowledge bank \ie codebook. 
Tokenization is widely used in various applications like image synthesis~\cite{vqvae, vqvae2}, text-to-image
generation~\cite{vqvae_text2img}, 2D human pose estimation~\cite{Geng23PCT}, and learning motion priors~\cite{t2mgpt, jiang2023motiongpt}.  
In the context of human pose estimation, tokenization
remains relatively unexplored, though it is widely used in human motion generation \cite{t2mgpt, tm2t}.
Of course, tokenized representations of images and language are widely used for many vision and language problems.
Our approach is novel in that it reformulates the regression problem as a pose token classification problem. It thus exploits tokenization to represent valid poses, effectively providing a pose prior.

\section{Camera/pose Bias}
\label{bias}

\begin{figure}[t]
  \centering  
  \includegraphics[width=\columnwidth]{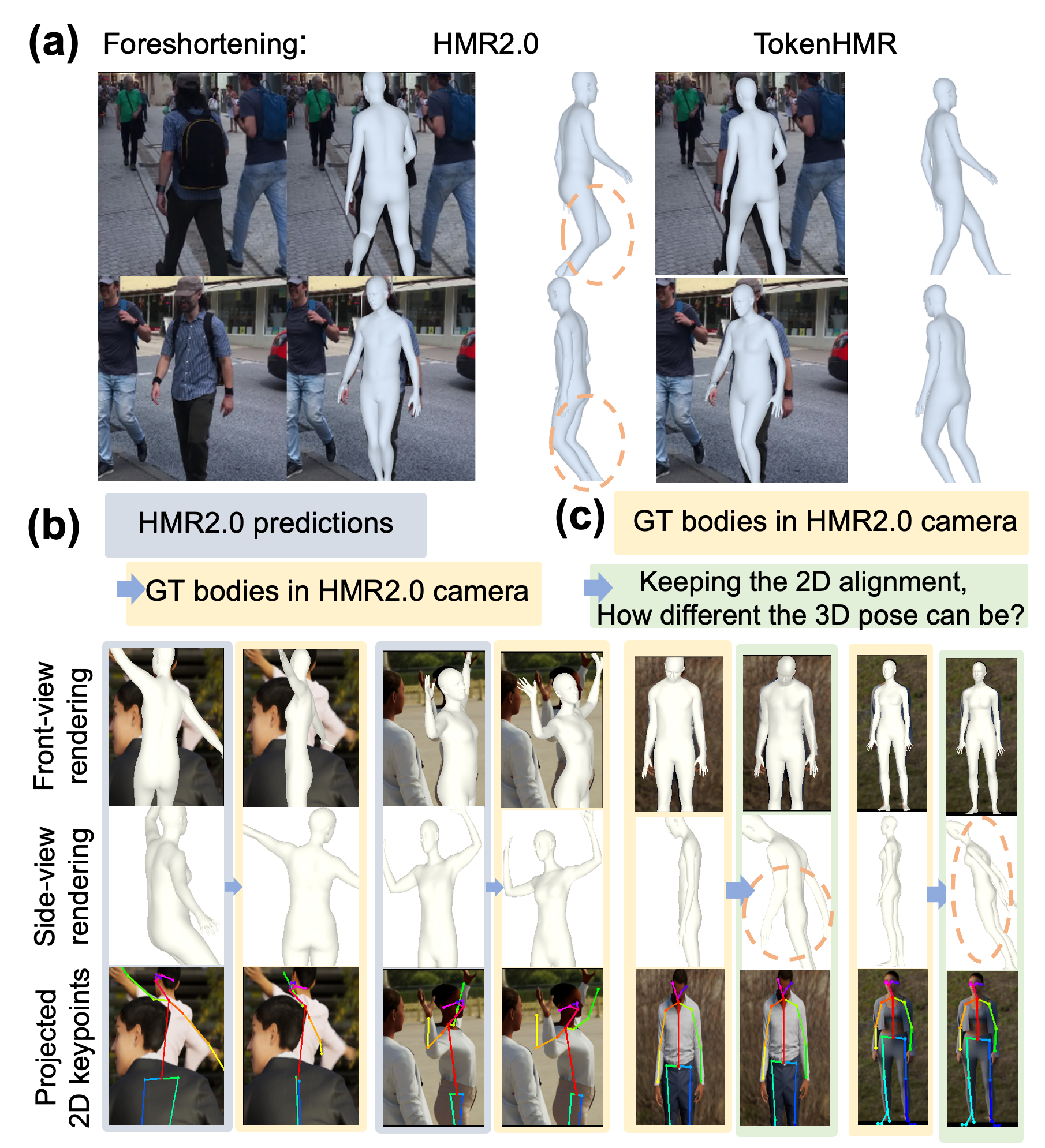}
  \vspace{-5mm}
  \caption{\textbf{Visualization of the camera/pose bias issues.} 
  \CR{(a) The lack of correct focal length means that foreshortened legs are estimated as bent by methods like HMR2.0.
  (b) Replacing the predicted body poses with ground truth reveals camera bias; (c) Maintaining 2D alignment, how wrong can the 3D poses be? See ~\cref{bias} for details.}}
  \vspace{-3mm}
  \label{fig:exp1}
\end{figure}

Methods that estimate 3D HPS typically try to satisfy two goals: accurate 3D pose and accurate alignment with 2D image features.
Unfortunately, we observe a trend in all experiments -- the better a method does on 2D error, the worse it does on 3D and vice versa.
The key reason for this is that current methods, including those tested here, do not estimate the camera intrinsic parameters (e.g.~focal length) or the camera extrinsics (rotation and translation).
Instead, current methods estimate the person in camera coordinates using scaled orthographic projection or perspective projection with fixed and incorrect camera parameters.
This results in a mismatch between the true 3D joints and their 2D projection.
Specifically, since photos are typically taken from roughly eye height, the legs are further away than the upper body. This causes them to be foreshortened.
Training models to minimize 2D error forces them to generate incorrect poses in 3D; this is illustrated in Fig.~\ref{fig:exp1} (a).
Pseudo ground truth (p-GT) for 2D pose datasets is obtained by minimizing the 2D error with problematic camera parameters.
Fully trusting such p-GT and pursuing accurate learning of such annotations will make the problem more prominent.
Notice how foreshortening makes the legs appear shorter in the image.
The only way to make a human body fit this is to bend the legs at the knees  or tilt the body in 3D, making the legs further away.
This produces unnatural or unstable poses. 

This is a fundamental issue with all current methods and one cannot get low error for both 3D and 2D without knowing the camera.
To numerically evaluate the impact of this mismatch, we employ \bedlam \cite{bedlam}, a synthetic dataset where both 3D and 2D data are known exactly along with the camera.
This removes any possible noise and allows us to see the effects of using the wrong camera on 2D projection error.
Specifically, as shown in Fig.~\ref{fig:exp1} (b), we take ground truth \bedlam bodies and project them into the image using the camera of \hmrTwo \cite{hmr2}.

We evaluate the effect of the incorrect camera in 2D using the standard measure of Percentage of Correct Keypoints (PCK), which we compute for a sequence from the \bedlam validation set.
The 3D bodies computed by HMR2.0b have errors of 0.78 on PCK0.5 and 0.88 on PCK1.0. 
In contrast, when we use the HMR2.0b camera with the {\em ground truth} 3D bodies, the PCK scores {\em decrease} to 0.66 on PCK0.5 and 0.86 on PCK1.0. 
Ideally, with a correct camera model, both PCK0.5 and PCK1.0 should reach 1.0. 
The fact that HMR2.0b  achieves lower error than the ground truth indicates that its output deviates from the true 3D pose and shape due to camera bias. This demonstrates that methods like HMR2.0b, while obtaining high PCK values, do so at the expense of 3D accuracy.
In summary, seeking high PCK values is counterproductive to 3D accuracy unless one has the correct camera model.

We further design experiments to explore how bad the 3D error can be  while maintaining good 2D alignment. 
We modify the loss function of SMPLify \cite{smplify} to keep the distance between predicted 2D keypoints and GT 2D keypoints $\boldsymbol{\gtTwoDJnts}$ close, while adding a new loss to {\em increase} the distance between predicted 3D keypoints $\boldsymbol{\threeDJnts}$ and real 3D keypoints $\boldsymbol{\gtThreeDJnts}$, as expressed in the following equation:
\begin{equation}
        w_{2D}||\Pi(\boldsymbol{\threeDJnts},\boldsymbol{\cam}) - \boldsymbol{\gtTwoDJnts}||_2 - w_{3D}||\boldsymbol{\threeDJnts} - \boldsymbol{\gtThreeDJnts}||_2 +
        m
\label{formula:triplet_loss}
\end{equation}
where $\Pi$ represents 3D-to-2D projection using HMR2.0's camera, $m=20$ is the margin value, $w_{2D}=4$ and $w_{3D}=40.5$ are scalar weights.
After 100 iterations of optimization, the Mean Per Joint Position Error (MPJPE) reaches 146mm. As shown in Fig.~\ref{fig:exp1} (c), the projected 3D pose can still maintain a high degree of 2D alignment even with significant errors in the depth direction. When optimized for 200 iterations, the MPJPE exceeds 300mm, and the error continues to increase with further optimization.

Since the field does not currently have a reliable way to estimate the camera parameters from a single image, 
below we explore the ability of our new methods (TALS and tokenization)  to help mitigate the issues caused by approximate camera models.
Figure \ref{fig:exp1} (a) compares results from \hmrTwo and \modelname.  Note that the effect of foreshortening has less impact on pose with \modelname.

\section{TokenHMR}
\label{sec:method}

\begin{figure*}[t]
  \centering  
  \includegraphics[width=1\textwidth]{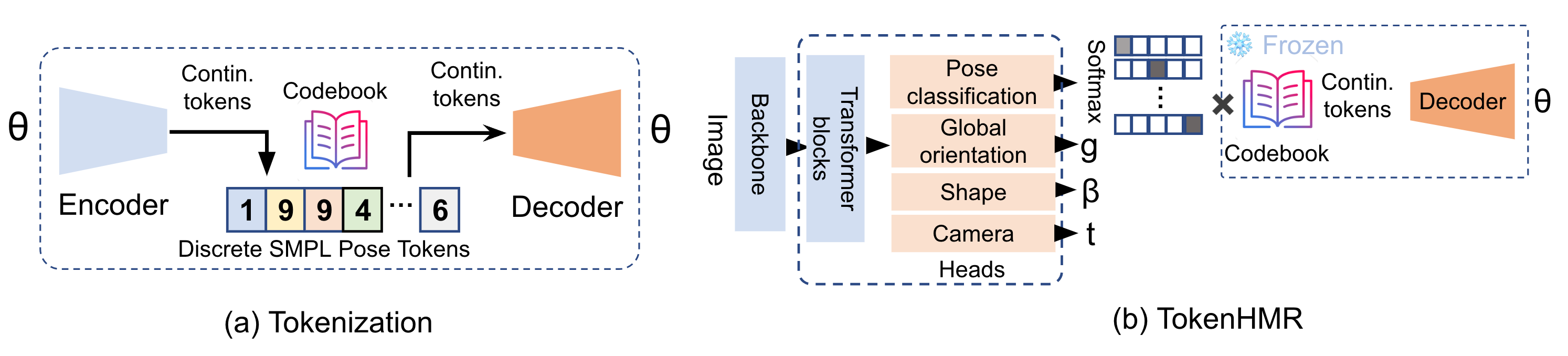}
  \vspace{-7mm}
  \caption{\textbf{Framework overview.} Our method has two stages. (a) In the tokenization step, the encoder learns to map continuous poses to discrete pose tokens and the decoder tries to reconstruct the original poses. 
  (b) To train \modelname, we replace regression with classification using the pre-trained decoder, which provides a ``vocabulary'' of valid poses.}
  \vspace{-3mm}
  \label{fig:framework}
\end{figure*}

\subsection{Preliminaries}
Our method, \modelname, takes an input image, $\image$, and outputs body pose, $\regPose$, shape, $\regShape$ and perspective camera, $\cam$.
We use \smpl~\cite{smpl}, a differentiable parametric body model. Its input parameters consist of pose, denoted by $\regPose \in \mathbb{R}^{72}$ and shape, denoted as $\regShape \in \mathbb{R}^{10}$. As output, it produces a body mesh, $\mesh$, and vertices, $\vertices \in \mathbb{R}^{NX3}$, where $N = 6890$ is the number of vertices. \threeD joints denoted as $\threeDJnts$, are derived through a linear combination of mesh vertices using a pre-trained joint regressor.

\subsection{\losscutFull: \losscutShort}
\label{sec:losscutAnalysis}

In Section~\ref{bias}, our analysis reveals a notable impediment to the effective learning using pseudo-ground-truth and 2D keypoints—camera/pose bias.
Despite this challenge, the scale provided by such annotations remains integral to achieving optimal generalization and robustness.
We assert that, when appropriately utilized, without over-fitting, these annotations significantly enhance the model's ability to robustly estimate pose.
A key insight emerges from our observations: establishing an effective threshold is imperative to discern the error levels that yield no additional benefit as a training signal. 
When the loss surpasses this threshold, conventional learning mechanisms guide pose estimation. 
Conversely, when the loss falls below this effective threshold, we minimize its impact to prevent over-fitting to the camera/pose bias.

To determine this effective threshold, we analyze the errors obtained using ground truth (GT) 3D poses and a standard (incorrect) camera model.
Again we leverage  the 3D GT in  \bedlam~\cite{bedlam}, 
this time to establish
effective thresholds for both \twoD keypoints and \smpl pseudo-ground-truth.
For 2D keypoints, we replace the predicted \smpl parameters with ground truth values from \bedlam to obtain the real 3D human body's 2D keypoint projections under the HMR2.0 camera.
We then calculate the mean L1 norm between these projections and the GT 2D keypoints, and use this as the threshold $\boldsymbol{\varepsilon_{\twoDJnts}}$ for 2D keypoint supervision. 
We normalize these 2D keypoints relative to image width and scale the values between -0.5 and 0.5 to mitigate scale-related variances.

Similarly, to establish the effective supervision threshold for \smpl p-GT, we conduct additional experiments.
With p-GT, we formulate the pose loss in terms of joint angle error.
To set appropriate thresholds on these errors, we evaluate the difference in joint angles between HMR2.0's predictions on \bedlam and the ground truth values for each joint in the \smpl model.
Specifically, we compute the mean geodesic distance\footnote{\url{https://rotations.berkeley.edu/geodesics-of-the-rotation-group-so3/}} 
between the 3D joint rotations on the manifold of rotations in SO(3) predicted by HMR2.0 and the ground-truth rotations in \bedlam. 
Please refer to Sec.~B.2 in \supmat for specific threshold values. 

After establishing the effective thresholds for 2D keypoints and \smpl p-GT, we introduce a new loss called \textit{\losscutFull (\losscutShort)}.
It scales down the loss only when it goes below the threshold. 
Specificially, the \textit{\losscutShort} loss terms for \pseudoGT pose and \twoD joints are defined as
\begin{equation}
    \mathcal{L}_{\boldsymbol{{\theta}_{pGT}}} = \begin{cases} 
      \| \boldsymbol{\regPose} - \boldsymbol{\gtPose} \|^2 & \text{if } \mathcal{L}_{\boldsymbol{{\theta}_{pGT}}} > \boldsymbol{\varepsilon_{\theta}} \\
      \alpha_{\theta} \cdot \| \boldsymbol{\regPose} - \boldsymbol{\gtPose} \|^2 & \text{otherwise}
   \end{cases}
\label{formula:tals3d}
\vspace{-0.4 em}
\end{equation}
\begin{equation}
    \mathcal{L}_{{J_{2D}}_{pGT}} = \begin{cases} 
      \vert \boldsymbol{\twoDJnts} - \boldsymbol{\gtTwoDJnts} \vert & \text{if } \mathcal{L}_{\boldsymbol{{J_{2D}}_{pGT}}} > \boldsymbol{\varepsilon_{\twoDJnts}} \\
      \alpha_{\twoDJnts} \cdot \vert \boldsymbol{\twoDJnts} - \boldsymbol{\gtTwoDJnts} \vert & \text{otherwise}
   \end{cases}
   \label{formula:tals2d}
\vspace{0.2 em}
\end{equation}
where $\alpha_{\twoDJnts}$ and $\alpha_{\theta}$ are small scalar multipliers and $\boldsymbol{\varepsilon_{\theta}}$ is the threshold calculated separately for each pose parameter and $\boldsymbol{\varepsilon_{\twoDJnts}}$ is the threshold for \twoD joints.

\subsection{Tokenization}
We use a \vqvae~\cite{vqvae}, which learns an encoding of 3D pose in a discrete representation. Specifically, we learn a discrete representation for \smpl body parameters, $\regPose = [\regPose_1, \regPose_2, \ldots, \regPose_{21}]$, where the $\regPose_i$ represent each joint's pose parameters in $\mathbb{R}^{6}$. The process involves encoding and decoding the pose parameters using an autoencoder architecture and a learnable \codebook, denoted as $\boldsymbol{\codebookBig} = \{\codebookSmall_k\}_{k=1}^K$, with each code $\codebookSmall_k \in \mathbb{R}^{d_c}$, where $d_c$ is the dimension of the codes. The overall architecture of the Pose VQ-VAE is illustrated in Fig.~\ref{fig:framework} (a). The encoder and decoder of the autoencoder are represented by $E$ and $D$, respectively. The encoder is responsible for generating discrete pose tokens, 
while the decoder reconstructs these tokens back to \smpl poses. The latent feature $z$ can be computed as $z = E(\regPose)$, resulting in $z = [z_1, z_2, \ldots, z_{M}]$, where $z_i \in \mathbb{R}^{d_c}$ and $M$ is the number of tokens. Each latent feature $z_i$ is quantized using the \codebook $\boldsymbol{\codebookBig}$ by finding the most similar code element, as expressed in the following equation:
\begin{equation}
\hat{z_i} =  \underset{\codebookSmall_k \in \boldsymbol{\codebookBig}}{\arg\min}\|z_i - \codebookSmall_k\|_2 .
\label{formula:quantization}
\end{equation}
In training the pose \tokenizer, we adopt a strategy similar to previous work~\cite{vqvae}, which involves three primary loss functions to optimize the \tokenizer: the reconstruction loss ($\mathcal{L_{RE}}$), the embedding loss ($\mathcal{L_{E}}$), and the commitment loss ($\mathcal{L_{C}}$). The overall loss ($\mathcal{L_{VQ}}$) is defined as
\begin{equation}
    \begin{split}
        \mathcal{L_{VQ}} 
        &= \lambda_{\mathcal{RE}}\mathcal{L_{RE}} + 
        \lambda_{\mathcal{E}}\mathcal{L_{E}} +
        \lambda_{\mathcal{C}}\mathcal{L_{C}}
        \\ 
        &= \lambda_{\mathcal{RE}}\mathcal{L_{RE}} + 
        \lambda_{\mathcal{E}}||\mathit{sg}[\boldsymbol{z}] - \boldsymbol{\embedding}||_2 + 
        \lambda_{\mathcal{C}}||\boldsymbol{z} - \mathit{sg}[\boldsymbol{\embedding}]||_2
    \end{split}
\label{formula:vqvae-loss}
\end{equation}
where $\mathit{sg}$ is the stop gradient operator, $\embedding$ is the embedding from the codebook and $\lambda_{\mathcal{RE}}$, $\lambda_{\mathcal{E}}$, $\lambda_{\mathcal{C}}$ are the hyper-parameters of for each term. For reconstruction, we use an $\mathcal{L}_1$ loss between the \groundtruth pose, $\gtPose$, and predicted pose, $\regPose$ and also on the error between the \smpl\/ \groundtruth \threeD joints, $\gtThreeDJnts$ and predicted joints, $\threeDJnts$. So, the $\mathcal{L_{RE}}$ loss is defined as
\begin{equation}
    \mathcal{L_{RE}} = \mathcal{L}_1(\boldsymbol{\gtPose}, \boldsymbol{\regPose}) + \mathcal{L}_1(\boldsymbol{\gtThreeDJnts}, \boldsymbol{\threeDJnts}).
\end{equation}
The original \vqvae suffers from \codebook collapse, \ie the \codebook is not fully utilized. Following prior work~\cite{t2mgpt}, we use the training strategy of exponential moving average (EMA) and codebook reset (Code Reset) for better utilization.

\subsection{Architecture}
Our architecture exploits the Vision Transformer (ViT)~\cite{vit}, similar to \hmrTwo~\cite{hmr2}. 
The input image, $\image$, is first transformed into input tokens, which are subsequently processed by the transformer to generate output tokens. These output tokens then undergo further processing in the transformer decoder. The transformer decoder has multi-head self-attention that cross-attends a zero input token with an image output token to get features from the transformer block. 
In contrast to HMR2.0 \cite{hmr2}, which employs three linear layers to map the features from transformer block to the \smpl pose, $\regPose$, shape, $\regShape$ and camera, $\cam$, we propose a novel approach. Our objective is to leverage a \tokenizer trained on a significant amount of motion capture (mocap) data, specifically focusing on body pose. To facilitate this, we partition the \smpl pose parameters into body pose and global orientation. We use separate linear layers to predict the global orientation and body pose from the \tokenizer. 

A straightforward integration of the \tokenizer would involve estimating the code index directly from the ViT transformer backbone and selecting embeddings, $\embedding$, based on the code index from the codebook, $\boldsymbol{\codebookBig}$. However, this poses a challenge as the process of selecting an embedding from the codebook, is non-differentiable. To address this issue, we adopt a logit-based approach. Instead of directly estimating the code index, we output logits, $\boldsymbol{\mathit{Q}}$ for each token. These logits are multiplied with the codebook, resulting in weighted embeddings. Thus, the approximated quantized feature $\bar{z} = [\bar{z}_1, \bar{z}_2, \ldots, \bar{z}_{M}]$ can be calculated as 
\begin{equation}
        \bar{z} = \sigma(\boldsymbol{\mathit{Q}}_{M \times K}) \times \boldsymbol{\codebookBig}_{K \times D} \approx \hat{z}
\end{equation}
where, $\boldsymbol{\mathit{Q}}$ are the logits estimated by the backbone, $\sigma$ is the softmax operation, $\boldsymbol{\codebookBig}$ is the pretrained codebook, $M$ is the number of tokens, $K$ is the number of entries in the codebook and $D$ is the dimension of each codebook entry.
The operation makes it differentiable. The obtained approximated quantized features, $\bar{z}$ are subsequently passed through the \tokenizer decoder. This process yields the final pose. 
In the training of \modelname, the learned \tokenizer decoder is frozen to take advantage of the 
prior it has learned from mocap data.

\subsection{Losses}

Following prior work~\cite{pare,hmr2}, we define losses on \twoD and \threeD joints and \smpl pose and shape parameters, \ie on $\boldsymbol{\twoDJnts}$, $\boldsymbol{\threeDJnts}$, $\boldsymbol{\regPose}$, $\boldsymbol{\regShape}$, respectively. However, following the analysis in ~\cref{sec:losscutAnalysis}, we treat data from \twoD and \threeD datasets differently. For \threeD \groundtruth datasets, we define the standard loss as
\begin{equation}
    \begin{split}
        \mathcal{L}_{GT} = \lambda_{\regPose}\mathcal{L}_{\regPose}(\boldsymbol{\regPose}, \boldsymbol{\gtPose}) + \lambda_{\regShape}\mathcal{L}_{\regShape}(\boldsymbol{\regShape}, \boldsymbol{\gtShape}) + \\
        \lambda_{3D}\mathcal{L}_{3D}(\boldsymbol{\threeDJnts}, \boldsymbol{\gtThreeDJnts}) + 
        \lambda_{2D}\mathcal{L}_{2D}(\boldsymbol{\twoDJnts}, \boldsymbol{\gtTwoDJnts})
    \end{split}
\vspace{-0.5em}
\end{equation}
where $\mathcal{L}_{\regShape}$ is a \smpl shape loss, 
$\mathcal{L}_{\threeDJnts}$ is the \threeD joint loss and  $\mathcal{L}_{\twoDJnts}$ is the joint re-projection loss. $\lambda_{\regShape}$, $\lambda_{3D}$ and $\lambda_{2D}$ are steering weights for each term. 
To learn from \smpl pseudo-ground-truth, we use \textit{\losscutFull (\losscutShort)} where we scale the loss based on the threshold computed in ~\cref{sec:losscutAnalysis}, outlined in Eqs.~\ref{formula:tals3d}  and~\ref{formula:tals2d}.
Thus, the total loss is defined as
\begin{equation}
    \mathcal{L}_{Total} = \mathcal{L}_{GT} + \mathcal{L}_{{\theta}_{pGT}} + \mathcal{L}_{{J_{2D}}_{pGT}} .
\end{equation}

\section{Experiments}
\subsection{Implementation Details}
\label{sec:implementation}

\begin{table*}[t]
    \centering
    \footnotesize
    \resizebox{0.80\textwidth}{!}{
    \begin{tabular}{l|l|ccc|ccc}
        \toprule
        Training & \multirow{2}{*}{Method} & \multicolumn{3}{c|}{EMDB~\cite{emdb}}& \multicolumn{3}{c}{3DPW~\cite{threedpw}} \\
        \cmidrule{3-8}
        Datasets& & MVE & MPJPE & PA-MPJPE & MVE & MPJPE & PA-MPJPE \\
        \midrule
        SD & HybrIK~\cite{hybrik} & 122.2 & 103.0 & 65.6 & 94.5 & 80.0 & 48.8  \\
        SD & CLIFF~\cite{cliff} & 122.9 & 103.1 & 68.8 & 81.2 & 69.0 & 43.0 \\
        SD &HMR2.0~\cite{hmr2} & 120.1 & 97.8 & 61.5 & 84.1 & 70.0 & 44.5 \\ %
        BL & BEDLAM-CLIFF~\cite{bedlam} & 113.2  & 97.1 &61.3  & 85.0 & 72.0 &  46.6  \\ %
        \CR{BL} & \CR{HMR2.0} & 106.6 & 90.7 & 51.3 & 88.4 & 72.2 & 45.1 \\
        \CR{BL} & \CR{TokenHMR} & 104.2 & 88.1 & 49.8 & 86.0 & 70.5 & 43.8 \\
        \cmidrule{1-8}
        SD + ITW &
        HMR2.0~\cite{hmr2} & 140.6 & 118.5 & 79.3 & 94.4 & 81.3 & 54.3 \\ %
        SD + ITW &
        TokenHMR & 124.4 & 102.4 & 67.5 & 88.1 & 76.2 & 49.3 \\
        \cmidrule{1-8}
        SD + ITW + BL &
        HMR2.0 & 120.7 & 99.3 & 62.8 & 88.4 & 77.4 & 47.4 \\
        SD + ITW + BL &
        HMR2.0 +  \losscutShort & 115.7 & 96.7 & 58.5 & 89.6 & 73.5 & 46.8 \\
        SD + ITW + BL &
        HMR2.0 + Token & 116.1 & 95.6 & 62.2 & 86.6 & 75.0 & 48.0 \\
        SD + ITW + BL &
        HMR2.0 + \losscutShort + \vposer \cite{smplx} & 116.8 & 97.9 & 56.4 & 87.1 & 73.7 & 45.7 \\
        SD + ITW + BL &
        TokenHMR & 109.4 & 91.7 & 55.6 & 84.6 & 71.0 & 44.3  \\ %
        \bottomrule
    \end{tabular}
    }
    \caption{3D human mesh and pose errors on the EMDB and 3DPW datasets. See text.} 
    \label{tab:comp_sota}
\end{table*}

\begin{table*}[t]
    \centering
    \footnotesize
    \resizebox{0.85\textwidth}{!}{
    \begin{tabular}{l|ccc|ccc}
        \toprule
        \multirow{2}{*}{Method} & \multicolumn{3}{c|}{Crop 30\%}& \multicolumn{3}{c}{Crop 50\%} \\
        \cmidrule{2-7}
        & MVE & MPJPE & PA-MPJPE & MVE & MPJPE & PA-MPJPE \\
        \midrule
        HMR2.0~\cite{hmr2} & 135.24 (+14.98) & 113.39 (+14.13)  & 70.68 (+7.86) & 166.71 (+46.45) & 137.88 (+38.59) &  90.30 (+27.48)  \\
        TokenHMR & 124.09 \textbf{(+14.71)} & 104.72 \textbf{(+13.01)} & 62.13 \textbf{(+6.52)} & 150.29 \textbf{(+40.91)} & 125.99 \textbf{(+34.28)} & 78.88 \textbf{(+23.27)} \\ 
        \bottomrule
    \end{tabular}
    }
    \caption{Impact of evenly cropping images at different ratios from the boundaries on the 3D HPS accuracy on the EMDB dataset. The numbers in (parentheses) indicate the changes in performance relative to the non-cropped scenario; smaller is better. All models compared here employ identical backbones and are trained on the same data.} 
    \label{tab:crop_study}
\end{table*}

\begin{table}[t]
	\centering
	\scriptsize
	\resizebox{\columnwidth}{!}{
		\begin{tabular}{l|c|c|c|c|c}
			\toprule
			& & \multicolumn{2}{c|}{ \amass \cite{amass} } & \multicolumn{2}{c}{\moyo \cite{moyo}}\\
			\midrule
			& Method & {MVE $\downarrow$} & {MPJPE $\downarrow$} & {MVE $\downarrow$} & {MPJPE $\downarrow$} \\
			\midrule
                \parbox[t]{1mm}{\multirow{3}{*}{\rotatebox[origin=c]{90}{CB}}}
                         & 1024 $\times$ 256     &    11.5      &     4.6   &      27.1    &      15.7     \\
                        & 2048 $\times$ 128     &     9.4      &     3.1    &      22.5    &      12.3     \\
                        & 2048 $\times$ 256     &     8.3      &     2.2    &      19.9    &      10.4     \\
                \midrule
                \parbox[t]{1mm}{\multirow{3}{*}{\rotatebox[origin=c]{90}{Tokens}}}
                        & 80     &      12.5      &     4.1    &      24.4    &      16.7     \\
                        & 160     &     8.3      &     2.2    &      19.9    &      10.4     \\
                        & 320     &     8.1      &     1.9    &      19.0    &      10.1     \\
                \midrule
                \parbox[t]{1mm}{\multirow{2}{*}{\rotatebox[origin=c]{90}{Noise}}}
                        & Yes     &     8.3      &     2.2    &      19.9    &      10.4     \\
                        & No     &     7.9      &     1.9    &      21.0    &      11.5     \\
                \midrule
                \multicolumn{2}{c|}{AMASS + $\text{MOYO}^\star$}     &     8.7    &    2.6  &    16.5 &  7.6    \\
			\bottomrule
		\end{tabular} 
	}
	\caption{
	            \cheading{Tokenizer Ablation}
	    All methods are trained on the standard training set of AMASS~\cite{amass} and evaluated on the test set of AMASS and validation set of MOYO~\cite{moyo} except the last $\text{row}^\star$, which is trained with the MOYO training set. The last model is used as the tokenizer in \modelname.
	}
        \vspace{-2mm}
	\label{tab:tokeniser_ablation}
\end{table}

\begin{figure*}[t]
  \centering  
  \includegraphics[width=1.\textwidth]{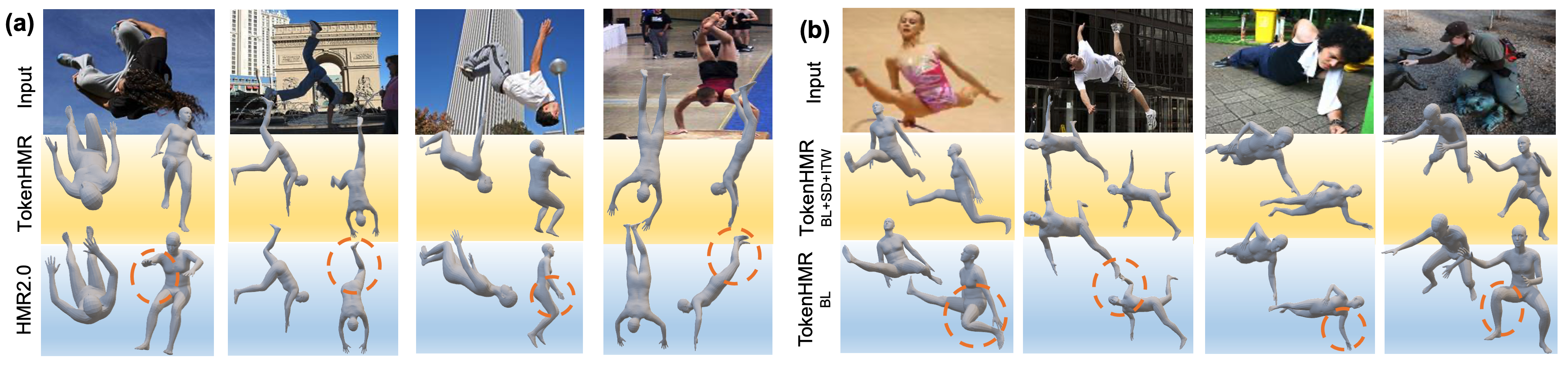}
  \vspace{-6mm}
  \caption{Qualitative comparisons on challenging poses from the LSP~\cite{lspet} dataset.  }
  \vspace{-2mm}
  \label{fig:truncation}
\end{figure*}

Training of \modelname involves two stages: first we train a tokenizer to learn discrete pose representations using AMASS~\cite{amass} and MOYO~\cite{moyo} mocap data. Then we use the pretrained decoder of the tokenizer as an additional head for regressing body pose. During the training of \modelname, the tokeniser is frozen to exploit the 
prior.

Our tokenizer architecture is inspired from T2M-GPT~\cite{t2mgpt} but instead of learning motion tokens \REFINE{of 3D joints} we learn pose tokens \REFINE{of SMPL pose parameters}. We use 1 ResNet~\cite{resnet} block and 4 1D convolutions both in the encoder and decoder. The steering weights $\lambda_{\mathcal{RE}}$, $\lambda_{\mathcal{E}}$, $\lambda_{\mathcal{C}}$ are set at $50.0, 1.0, 1.0$, respectively. The model is trained for $150K$ iterations with batch size of 256 and learning rate of $2e^{-4}$. To train a robust model, we augment random joints with noise starting from $1e^{-3}$, which we progressively increase after every $5K$ iterations. We choose the best tokenizer model containing $160$ tokens and codebook of size $2048 \times 256$ for \modelname based on reconstruction error on the validation set.

For \modelname, we use ViT-H/16~\cite{vit} as the backbone and standard transformer decoder~\cite{attention} following HMR2.0~\cite{hmr2}. We use 4 separate linear layers to map the features of size $1024$ from the transformer decoder to the global orientation, hand pose, and body shape of \smpl and one for the camera. However, for body pose, we process the $1024$ features through 4 blocks of linear layers, each containing 2 MLPs and an GELU activation function~\cite{gelu}. This gives the final logits, $\boldsymbol{\mathit{Q}}$, of size $160 \times 2048$ for multiplication with a codebook of size $2048 \times 256$, which results in approximate quantized features, $\bar{z}$; see Eq.~\ref{formula:quantization}. We use ViTPose~\cite{vitpose} as the pretrained backbone. We train for $100K$ iterations on 4 Nvidia RTX 6000 GPUs with a batch size of 256 and learning rate of $1e^{-5}$ for about one day. The steering weights, $\lambda_{\regPose}$, $\lambda_{\regShape}$, $\lambda_{\twoDJnts}$, $\lambda_{\threeDJnts}$ are set to $1e^{-3}$,  $5e^{-4}$,  $1e^{-2}$,  $5e^{-2}$, respectively. 
The loss weights of \textit{\losscutShort} is set to 1\% for both pose $\alpha_{\theta}$ and \twoD keypoints $\alpha_{\twoDJnts}$.

\textbf{Training Data:}
For training the tokenizer, we use the standard training split of AMASS~\cite{amass} and the training data of MOYO~\cite{moyo}. For more details on data preparation of training, please refer to \supmat
Following the prior methods~\cite{hmr2, pare, spin}, we use standard datasets (\textit{SD}) for training which include Human3.6M~\cite{human36m}, MPI-INF-3DHP~\cite{mpi-inf-3dhp}, COCO~\cite{coco}, and MPII~\cite{mpii}.
Additionally, like HMR2.0b, we also use in-the-wild 2D datasets (\textit{ITW}) like InstaVariety~\cite{humanMotionKanazawa19}, AVA~\cite{ava}, and AI Challenger~\cite{ai_challenger} datasets and their p-GT for training. 
We also include BEDLAM (\textit{BL})~\cite{bedlam}, a synthetic dataset with accurate ground-truth 3D data. For a fair comparison, we re-train HMR2.0b using a combination of  the SD, ITW, and BL datasets. We choose HMR2.0b as a baseline model since the code is open-source and we can reproduce the results.

\textbf{Evaluation and Metrics:}
For the tokeniser accuracy, we report the Mean Vertex Error (MVE) and Mean Per Joint Position Error (MPJPE) and evaluate on the standard test split of AMASS and validation set of MOYO.
For TokenHMR, we report the Mean Vertex Error (MVE), Mean Per Joint Position Error (MPJPE), and Procrustes-Aligned Mean Per Joint Position Error (PA-MPJPE) between the predictions and the ground-truth. We evaluate on the test set of the 3DPW~\cite{threedpw} and EMDB~\cite{emdb} datasets. The former is a standard 3D dataset and the latter is a recently released and more challenging dataset with varying camera motions and varied 3D poses.

\subsection{How to Alleviate the 3D Degradation Problem?}

Table \ref{tab:comp_sota} shows the performance of the HMR2.0 model trained solely with the \textit{SD} dataset and its performance when trained with both \textit{SD} and \textit{ITW} datasets. We observe a significant decrease (over 17\%) in 3D accuracy on the EMDB dataset upon the inclusion of the \textit{ITW} data.
At the same time, according to the Table 2 in HMR2.0~\cite{hmr2} paper,  the \textit{ITW} data improves the method's 2D performance.
A straightforward approach to counter this trend could be to integrate more data with precise 3D annotations, such as BEDLAM \cite{bedlam}. 
Yet, as Table \ref{tab:comp_sota} reveals, even with the inclusion of BEDLAM, HMR2.0 (\textit{SD}+\textit{ITW}+\textit{BL}) still suffers from noticeable 3D metric degradation. This observation forms our baseline for further investigations.

Employing our novel loss formulation (\textit{\losscutShort}) results in notable performance improvement on both the EMDB and 3DPW datasets, indicating its effectiveness in preventing overfitting to noisy \pseudoGT data. While \textit{\losscutShort} yields improvements, we delve deeper into exploring pose priors to compensate for the diminished supervision. Our evaluation of VPoser \cite{smplx}, a prevalent VAE prior in HPS, yields only marginal improvements, suggesting the need for a more robust alternative. 
Our VQ-VAE-based pose tokenization approach offers greater improvement.  
\modelname significantly outperforms HMR2.0, with improvements of 9\% in MVE, 7.6\% in MPJPE, and 11.5\% in PA-MPJPE on EMDB. Consistent trends are also observed on 3DPW.

\subsection{How does the Token-based Prior Help?}
Beyond facilitating more effective learning as we discussed above, we also examine the efficacy of our discrete token-based prior in scenarios with ambiguous image information, such as truncation.
We evaluate our method under varying degrees of image cropping on the EMDB dataset. Specifically, we crop 30\% and 50\% from the image boundaries.
As shown in Table \ref{tab:crop_study}, compared with HMR2.0 \cite{hmr2}, the performance of our approach decreases less in the challenging truncation settings (50\% v.s.~30\%).
Furthermore, the qualitative outcomes, illustrated in Fig.~\ref{fig:truncation}, underscore the robustness of the prior embedded within our token-based pose representation. This robustness is crucial for handling real-world scenarios where image truncation is common.

\subsection{Ablation Study of Tokenizer}
Table \ref{tab:tokeniser_ablation} presents our ablation study on different tokenizer design choices using AMASS's standard test set and MOYO's validation set. To understand the impact of the design choices on out-of-distribution MOYO data, we train solely with AMASS and conduct various ablations. 
The final tokenizer model (last row in Table~\ref{tab:tokeniser_ablation}), however, is used in  \modelname and is trained on both the AMASS and MOYO datasets. 
Our findings indicate that the number of codebook entries has a more significant impact than code dimensions. Although the number of tokens is crucial for an accurate representation, we observe a performance plateau, opting for 160 tokens in our final model. 
This number strikes a balance between network size and reconstruction accuracy for \modelname. Random augmentation of pose parameters with noise builds a more robust tokenizer, slightly reducing performance for in-distribution data but beneficially impacting OOD data.

\section{Conclusion}
\label{sec:conclusions}

In this paper, we presented a novel approach to 3D human pose estimation from single images.
We begin by identifying and quantifying the problem caused by using a 2D keypoint loss with an incorrect camera model.
This leads to a fundamental tradeoff for current methods -- either have high 3D accuracy or 2D accuracy, but not both.
Our method, \modelname, addresses this problem with two contributions that can easily be used by other methods.
\modelname adopts a new paradigm for pose estimation based on regressing a discrete tokenized representation of human pose.
We combine this with a new loss, \textit{\losscutShort}, which mitigates some of the bias caused by the camera projection error, and biased p-GT, while still allowing the use of in-the-wild training data.
Our experiments on the EMDB and 3DPW datasets demonstrate that \modelname significantly outperforms existing models like HMR2.0 in terms of 3D accuracy, even with siginficant occlusion.

{
\small
\section{Acknowledgement}
We sincerely thank the department of Perceiving Systems and ML team of Meshcapade GmbH for insightful discussions and feedback. We thank the International Max Planck Research School for Intelligent Systems (IMPRS-IS) for supporting Sai Kumar Dwivedi. 
We thank Meshcapade GmbH for supporting Yu Sun and providing GPU resources. 
This work was partially supported by the German Federal Ministry of Education and Research
(BMBF): Tübingen AI Center, FKZ: 01IS18039B.
\textbf{Disclosure:} \href{https://files.is.tue.mpg.de/black/CoI\_CVPR\_2024.txt}{
https://files.is.tue.mpg.de/black/CoI\_CVPR\_2024.txt}

}

{
    \small
    \bibliographystyle{config/ieeenat_fullname}
    \bibliography{config/main}
}

\clearpage

\appendix

\setcounter{figure}{0} \renewcommand{\thefigure}{S.\arabic{figure}}
\setcounter{table}{0} \renewcommand{\thetable}{S.\arabic{table}}

\clearpage
\setcounter{page}{1}
\maketitlesupplementary

\section{Introduction}

In this supplemental document, we provide more implementation details and discuss limitations of TokenHMR. 
Please refer to the \textbf{supplemental video} for a brief review of the paper and more qualitative results.

\section{More Implementation Details}

\subsection{Data Preparation for Tokenizer}
For pose tokenization, we use {21} body pose parameters following Vposer~\cite{smplx}. As shown in Tab. 3 of main paper, we evaluate our tokenization in two settings: in-distribution and out-of-distribution. For in-distribution, we train on the training set of AMASS~\cite{amass} and evaluate on the test set of AMASS. To show the efficacy of tokenization, we also evaluate on an out-of-distribution yoga dataset, MOYO~\cite{moyo}. For training, we use the following datasets: \small{\texttt{\{CMU, KIT, BMLrub, DanceDB, BMLmovi, EyesJapan, BMLhandball, TotalCapture, EKUT, ACCAD, TCDHands, MPI-Limits\}}} with a weighting of \{0.14, 0.14, 0.14, 0.06, 0.06, 0.06, 0.06, 0.06, 0.04, 0.04, 0.04, 0.16\}, respectively.

\subsection{Joint-wise Thresholds for TALS}
\begin{table}[ht]
    \centering
    \footnotesize
    \begin{tabular}{l|c|l|c}
        \toprule
        2D Joints & Threshold & SMPL Joints & Threshold \\
        \midrule
        OP Nose & 0.00850& Pelvis & 0.46\\
        OP Neck & 0.00649& LHip & 0.22\\
        OP RShoulder & 0.00748& RHip & 0.21\\
        OP RElbow & 0.01103& Spine & 0.15\\
        OP RWrist & 0.01356& LKnee & 0.33\\
        OP LShoulder & 0.00742& RKnee & 0.30\\
        OP LElbow & 0.01097& Thorax & 0.17\\
        OP LWrist & 0.01414& LAnkle & 0.20\\
        OP MidHip & 0.00974& RAnkle & 0.27\\
        OP RHip & 0.01127& Thorax & 0.12\\
        OP RKnee & 0.01663& LToe & 0.29\\
        OP RAnkle & 0.00565& RToe & 0.28\\
        OP LHip & 0.01126& Neck & 0.24\\
        OP LKnee & 0.01616& LCollar & 0.26\\
        OP LAnkle & 0.00533& RCollar & 0.26\\
        OP REye & 0.00830& Jaw & 0.28\\
        OP LEye & 0.00831& LShoulder & 0.29\\
        OP REar & 0.00737& RShoulder & 0.32\\
        OP LEar & 0.00743& LElbow & 0.35\\
        OP LBigToe & 0.00544& RElbow & 0.35\\
        OP LSmallToe & 0.00551& LWrist & 0.62\\
        OP LHeel & 0.00536& RWrist & 0.59\\
        OP RBigToe & 0.00565& LHand & 0.20\\
        OP RSmallToe & 0.00582& RHand & 0.20\\
        OP RHeel & 0.00573&  & \\
        LSP RAnkle & 0.00554&  & \\
        LSP RKnee & 0.01515&  & \\
        LSP RHip & 0.00986&  & \\
        LSP LHip & 0.00998&  & \\
        LSP LKnee & 0.01520&  & \\
        LSP LAnkle & 0.00511&  & \\
        LSP RWrist & 0.01288&  & \\
        LSP RElbow & 0.01106&  & \\
        LSP RShoulder & 0.00711&  & \\
        LSP LShoulder & 0.00710&  & \\
        LSP LElbow & 0.01092&  & \\
        LSP LWrist & 0.01388&  & \\
        LSP Neck & 0.00648&  & \\
        LSP Head Top & 0.00766&  & \\
        MPII Pelvis & 0.00931&  & \\
        MPII Thorax & 0.00647&  & \\
        H36M Spine & 0.00677&  & \\
        H36M Jaw & 0.00744&  & \\
        H36M Head & 0.00752&  & \\
        \bottomrule
    \end{tabular}
    \caption{Thresholds for 44 2D joints and 24 SMPL joints. 2D joint names start with the skeleton origin, where OP stands for OpenPose~\cite{openpose}. LSP~\cite{lspet}, MPII~\cite{mpii}, and H36M~\cite{human36m} are the datasets.} 
    \label{tab:joint_threshold}
\end{table}

To establish effective joint-wise thresholds for TALS (Sec. 4.2), we conducted a detailed statistical analysis on the 20221018\_3\-8\_250\_batch01hand\_6fps validation subset of the BEDLAM \cite{bedlam} dataset, encompassing over 34k samples of diverse human 3D pose, shape, and camera perspectives. 
Table \ref{tab:joint_threshold} presents the threshold distances for each joint used by TALS.

\CR{
\subsection{Augmentations}
Data augmentation plays a pivotal role in enhancing the robustness and generalization capabilities of \hps regressors.
Hence, following HMR2.0, we perform various augmentations. These include random translations in both \textit{x} and \textit{y} directions with a factor of $0.02$, scaling with a factor of $0.3$ and rotations with $30$ degrees. Other augmentations include horizontal flipping and color rescaling. We observe that extreme cropping \ie removing part of the human body limb in random also improves the robustness to occlusion.
}

\CR{
\section{Discussion}

\subsection{Pose Space Analysis}
We analyse the pose space by evaluating reconstruction of OOD poses that are not present in the training set. 
We do this by training on AMASS and testing on MOYO.
The qualitative result is shown in Fig.~\ref{fig:tsne_plot} which shows good generalization to the out-of-distribution yoga poses from MOYO~\cite{moyo}.
In contrast, we find that noisy test poses are not well recovered.

\subsection{TALS loss vs Filtering Strategy}
Similar to HMR2.0, we employ filtering strategies to ensure high-quality 2D image alignment of the p-GT. 
Filtering strategies, however, are ``all or nothing"; i.e.~data samples are either rejected or considered.
Our \losscutShort loss is different in that it uses all the filtered pseudo-ground-truth samples up to a threshold, after which the supervision is scaled down.
This goes beyond standard filtering and data cleaning pipelines.
}

\section{Limitation Discussion}

\subsection{Poor 2D Alignment under Weak-perspective Camera Model}

The experimental analysis in Sec.~3 shows that using existing flawed camera projection models results in overfitting to 2D keypoints and that this leads to learning biased poses.  %
To avoid this issue, we design a lenient TALS supervision training strategy and incorporate prior knowledge through our token-based pose representation. 
As shown in Fig.~\ref{fig:failures} a), with the combination of loose 2D supervision using TALS and built-in prior in representation, \modelname is able to estimate reasonable 3D poses but these do not always align well in 2D image when there is foreshortening.
As expected under the weak-perspective camera model, the more obvious the perspective distortion, the worse the 2D alignment.

\subsection{Failure Cases}
In this work, we introduce TokenHMR to reduce camera/pose bias and alleviate the ambiguity with a tokenized pose prior.
However, TokenHMR still has some limitations that could be further explored in future work.

As shown in Fig.~\ref{fig:failures} b), foreshortening remains challenging without a better camera model. %
In cases like Fig.~\ref{fig:failures} c), the global orientation is ambiguous when only considering body cues. We may need to exploit more cues from the face and the feet to determine the correct global orientation. Future work could try to extend TokenHMR to full-body pose estimation (i.e.~SMPL-X) to address this issue.

\begin{figure}[!t]
    \centerline{\includegraphics[width=\columnwidth]{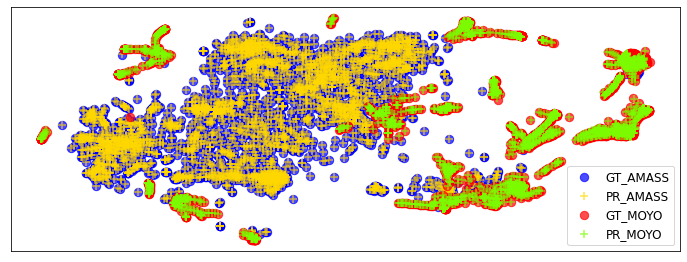}}
    \caption{
       \CR{t-SNE visualization of {\em unseen poses} (3D body joints) reconstructed by our tokenizer trained on AMASS only. 
       We are able to reconstruct the out-of-distribution Yoga poses from MOYO. GT is ground-truth poses and PR is predicted poses.
       }
    }
    \label{fig:tsne_plot}
    \vspace{-0.5 em}
\end{figure}

\begin{figure}[!tbh]
  \centering  
  \includegraphics[width=.5\textwidth]{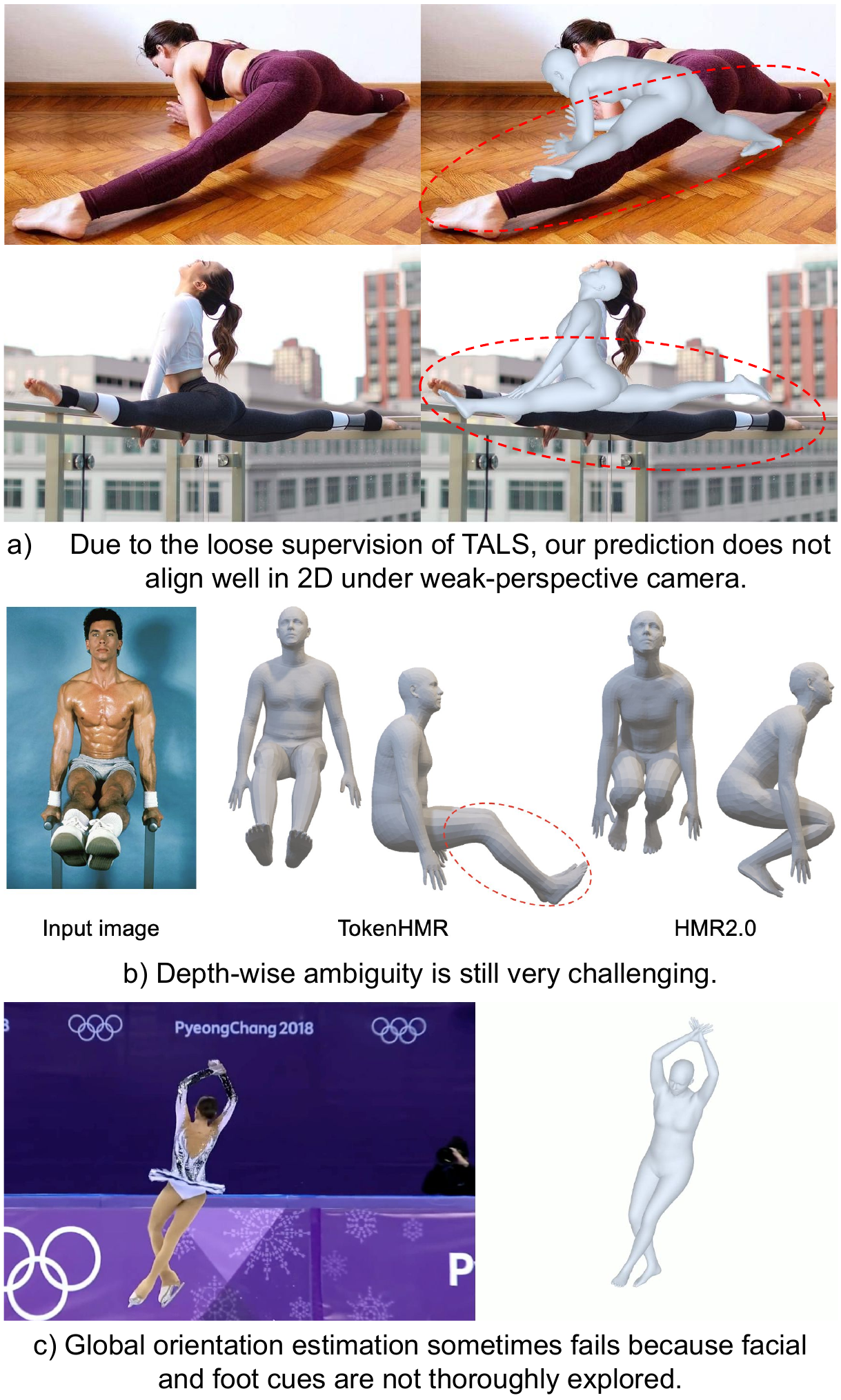}
  \caption{2D alignment problem and failure cases. }
  \vspace{-2mm}
  \label{fig:failures}
\end{figure}

\section{Future Work}
Future work should, obviously, address the camera projection problem directly by recovering more accurate camera estimates.
Even with such improvements, we anticipate that the token representation retains value as it consistently improves performance across varied test scenarios.
A promising next step is to extend the tokenization over time. 
Recent work on generating human motion from text exploits tokenized representations of human motions \cite{tevet2023human}.
Looking further ahead, an intriguing direction for future research involves exploring the application of our token-based pose representation with Large Language Models (LLMs). The discrete, robust nature of our pose tokens, designed for 3D human pose estimation, presents an opportunity to bridge the gap between computer vision and natural language processing.

\end{document}